\pdfoutput=1

\documentclass[11pt]{article}

\usepackage{acl22}

\usepackage{times}
\usepackage{latexsym}

\usepackage[T1]{fontenc}

\usepackage[utf8]{inputenc}

\usepackage{microtype}

\usepackage{soul}
\usepackage{url}
\usepackage[utf8]{inputenc}
\usepackage{graphicx}
\usepackage{amsmath}
\usepackage{amsthm}
\usepackage{booktabs}
\usepackage{algorithm}
\usepackage{algorithmic}
\title{Continuous Decomposition of Granularity for Neural Paraphrase Generation}

\author{
    Xiaodong Gu\textsuperscript{\rm 1},
    Zhaowei Zhang\textsuperscript{\rm 1},
    Sang-Woo Lee\textsuperscript{\rm 2},
    Kang Min Yoo\textsuperscript{\rm 2},
    Jung-Woo Ha\textsuperscript{\rm 2}\\
    $^1$ School of Software, Shanghai Jiao Tong University\\
    $^2$ NAVER AI Lab \\
    \{xiaodong.gu,andy\_zhangzw\}@sjtu.edu.cn\\
    \{sang.woo.lee,kangmin.yoo,jungwoo.ha\}@navercorp.com,
}

\usepackage{times}  
\usepackage{helvet}  
\usepackage{courier}  
\usepackage{subfig}
\usepackage{caption} 
\DeclareCaptionStyle{ruled}{labelfont=normalfont,labelsep=colon,strut=off} 
\usepackage{bibentry}

\usepackage{times}
\usepackage{latexsym}

\usepackage{microtype}
\usepackage{amsmath}
\usepackage{amsfonts,amssymb}

\usepackage{color}

\usepackage{xspace}
\newcommand{\approach}{C-DNPG\xspace}

\date{}

\begin{document}

\maketitle

\begin{abstract}
  While Transformers have had significant success in paragraph generation, they treat sentences as linear sequences of tokens and often neglect their hierarchical information.
  Prior work has shown that decomposing the levels of granularity~(e.g., word, phrase, or sentence) for input tokens has produced substantial improvements, suggesting the possibility of enhancing Transformers via more fine-grained modeling of granularity. 
  In this work, we propose continuous decomposition of granularity for neural paraphrase generation (\approach). In order to efficiently incorporate granularity into sentence encoding, \approach introduces a granularity-aware attention (GA-Attention) mechanism which extends the multi-head self-attention with:
 1) a granularity head that automatically infers the hierarchical structure of a sentence by neurally estimating the granularity level of each input token; and 2) two novel attention masks, namely, \emph{granularity resonance} and \emph{granularity scope}, to efficiently encode granularity into attention.
Experiments on two benchmarks, including Quora question pairs and Twitter URLs have shown that \approach outperforms baseline models by a remarkable margin and achieves the state-of-the-art results in terms of many metrics. Qualitative analysis reveals that \approach indeed captures fine-grained levels of granularity with effectiveness.
\end{abstract}

\section{Introduction}

With the continued success in NLP tasks~\cite{vaswani2017attention}, Transformer has been the mainstream neural architecture for paraphrase generation~\cite{li2019dnpg,kazemnejad2020paraphrase,ijcai2021-525,hosking2022hrq,goyal2020sow}.
The core component of Transformer is the self-attention network (SAN)~\cite{vaswani2017attention} which computes sentence representations at each position by baking representations over all other positions in a parallel way. 
Despite their effectiveness, Transformer has been shown to be limited in structure modeling, that is, they process disperse words in a flat and uniform way without explicit modeling of the hierarchical structures~\cite{Raganato2018san,hao2019multi,li2020transformer}.

\begin{table}
\centering
\small
\begin{tabular}{c|l}
\toprule
\bf Text  & What is the reason for World War II ?     \\
\bottomrule
decomposition 1  & \textcolor{blue}{What is} \textcolor{orange}{the reason for World War II} ?     \\
\hline
decomposition 2 & \textcolor{blue}{What is the} \textcolor{orange}{reason} \textcolor{blue}{for} \textcolor{orange}{World War II} ? \\
\hline
decomposition 3 & \textcolor{blue}{What is the reason for} \textcolor{orange}{World War II} ? \\
\hline
decomposition 4 & \textcolor{blue}{What is the reason for World War} \textcolor{orange}{II} ? \\
\hline
decomposition 5 & \textcolor{blue}{What is the reason for World War II} ? \\
\bottomrule
\end{tabular}
\begin{tabular}{c}
  $\Downarrow$\\
 \toprule
  \textbf{Levels of granularity} (marked as superscripts): \\
  \toprule
  What\textcolor{red}{$^1$} is\textcolor{red}{$^1$} the\textcolor{red}{$^2$} reason\textcolor{red}{$^3$} of\textcolor{red}{$^2$} World\textcolor{red}{$^4$} War\textcolor{red}{$^4$} II\textcolor{red}{$^5$} ? \\
  \bottomrule
\end{tabular}
\caption{A motivation example of multi-granularity text decomposition. The given sentence can be decomposed according to 5 increasing levels of granularity, each corresponding to a partition of the sentence into a template (\textcolor{blue}{blue}) at a specific level together with details (\textcolor{orange}{orange}). Each row with colored text denotes a level of granularity where the blue words are in the sentence level (templates) and the remaining words are in the phrase level (details).
The bottom half shows the level of granularity for each word according to the decomposition. We use integer numbers to indicate the extent of the granularity: the greater the number, the more detailed the word is. 
}
\label{motivateexample}
\end{table}

One potential route towards addressing this issues is multi-granularity text modeling (illustrated in Table~\ref{motivateexample}) – which decomposes texts into multiple levels of granularity such as words, phrase and sentence~\cite{li2019dnpg,wiseman2018learning,hao2019multi}. 
For example, Li et al. (2019) attempts to achieve this via \emph{decomposable neural paraphrase generator} (DNPG). DNPG decomposes paraphrase generation by Transformers into two levels of granularity: phrase-level (details) and sentence-level (templates). This allows a more flexible and controllable generating process. Paraphrases can be generated through rephrasing the templates while copying the detailed phrases. The decomposition is realized using a granularity separator, multiple encoder-decoder pairs, and an aggregator that summarizes results from all granularity levels.

Despite showing promising results, DNPG only captures a discrete (coarser-grained) decomposition of granularity, which restricts the capacity in representing more fine-grained semantic hierarchy. 
Furthermore, DNPG models the different levels of granularity using multiple encoders and decoders. That amounts to training multiple Transformers. The computational cost increases greatly as the number of granularity levels grows.

In this paper, we present \approach (stands for \emph{continuous decomposition of granularity for paraphrase generation}), a simple, fine-grained, and seamlessly integrated model for granularity-aware paragraph representation. \approach extends the vanilla attention network with a granularity head, which neurally estimates a continuous level of granularity for each token. In order to efficiently encode granularity into Transformers, \approach adjusts the original self-attention weights using two novel attention masks: 
1) a \emph{granularity-resonance mask} which encourages attentions to exist between tokens with similar granularity; 
and 2) a \emph{granularity scope mask} which encourages a small attention scope for lower-level (words or phrases) tokens. 
The granularity-aware attention mechanism provides a continuous modeling of sentence granularity and can be seamlessly integrated into the vanilla Transformer as the basic processing cell. 

We evaluate the proposed \approach on two commonly-used benchmarks, including the Quora question pairs and Twitter URL paraphrasing. Experimental results show that \approach remarkably outperforms baseline models on both benchmarks and achieve the state-of-the-art results in many metrics. Qualitative study confirms the ability of the proposed approach in modeling fine-grained granularity.

Our contributions can be summarized as follows:
\begin{itemize}
    \vspace{-0.2\baselineskip}
    \item We present a novel granularity-aware attention mechanism 
    which supports a fine-grained decomposition of granularity for input tokens and hence yields a \textbf{continuous modeling of granularity} for natural language sentences.
    \vspace{-0.3\baselineskip}
    \item The proposed granularity-aware attention network can be \textbf{seamlessly integrated} into the Transformer for granularity-aware paraphrase generation. 
    \vspace{-0.3\baselineskip}
    \item We conduct extensive evaluations of our methods on two popular paraphrase generation benchmarks and show that \approach remarkably outperforms previous works in terms of quantitative and qualitative results. We release all data and code at \verb|https://github.com/guxd/C-DNPG|.
\end{itemize}

\section{Background}
Our approach is extended based on the Transformer and self-attention networks. We begin by introducing the background of these techniques.
\subsection{Self-Attention Networks}
The attention mechanism is a function which maps a query vector to a set of key-value vector pairs and summarizes an output vector as a weighted sum of the value vectors. The weight assigned to each value is computed using the query and key vectors. A typical attention function, for example, is the scaled dot-product attention~\cite{vaswani2017attention}:
\begin{equation}
   \begin{split}
      &\text{Attention} (\mathbf{Q}, \mathbf{K}, \mathbf{V}) = \mathbf{A}^T\mathbf{V},\\
      &\mathbf{A} = \text{softmax}(\mathbf{QK}^T/\sqrt{d_k})
   \end{split}
\end{equation}
where {$\mathbf{Q}$, $\mathbf{K}$, $\mathbf{V}$}$\in\mathbb{R}^{L\times d}$ represent the query, key and value vectors, respectively. $\mathbf{A}\in[0,1]^{L\times L}$ is the attention score matrix with each $\mathbf{A}_{ij} = \mathbf{q}_i^T\mathbf{k}_j$; $d_k$ denotes the dimension size of the key vector.

In particular, the \emph{self-attention network} (SAN) is a special attention mechanism that computes the attention function over a single sequence. 
For a given sequence represented as a list of hidden states~$\mathbf{H}$ = [$\mathbf{h}_1,\ldots,\mathbf{h}_N$]$\in$ $\mathbb{R}^{N\times d}$, the self-attention network computes representations of the sequence by relating hidden states at different positions~\cite{vaswani2017attention}:
\begin{equation}
  \begin{split}
      & \mathbf{Q}, ~\mathbf{K}, \mathbf{V} = \mathbf{W}^Q\mathbf{H}, \mathbf{W}^K\mathbf{H}, \mathbf{W}^V\mathbf{H} \\
      & \mbox{SelfAttn}(\mathbf{H}) = \mbox{Attention}(\mathbf{Q}, \mathbf{K}, \mathbf{V})
  \end{split}
\end{equation}
where $\mathbf{W}^q$, $\mathbf{W}^k$, and $\mathbf{W}^v$ are parameters to transform the input representation $\mathbf{H}$ to the query, key, and value respectively.
Self-attention produces an abstraction and summary of a sequence in the hidden space and outputs the transformed hidden states~\cite{vaswani2017attention}.

\subsection{Transformer}
Transformer is an encoder-decoder model that is built upon the self-attention networks. It has been the common architectural choice for modeling paraphrase generation~\cite{li2019dnpg,ijcai2021-525,kazemnejad2020paraphrase}. Transformer encodes a source sequence into hidden vectors and then generates a target sequence conditioned on the encoded vectors. 
Both the encoder and the decoder are composed of a stack of $N$ identical layers, with each consists of a multi-head self-attention network followed by a position-wise fully connected network. 
Formally, the procedure of learning sequence representations through Transformer can be formulated as follows:
\begin{equation}
  \begin{bmatrix}
       \mathbf{\bar H}^l = \mbox{LN}(\text{SelfAttn}(\mathbf{H}^{l-1}) + \mathbf{H}^{l-1}) \\
       \mathbf{H}^l = \mbox{LN}(\text{FFN}^l(\mathbf{\bar H}^l) + \mathbf{\bar H}^l)
  \end{bmatrix}_L
\end{equation}
 where SelfAttn$(.)$ denotes the multi-head self-attention network which performs the attention function over $\mathbf{H}^{l-1}$, the output hidden states of the $l$-1$^{st}$ layer; FFN and LN stand for the position-wise fully connected layer and the layer normalization, respectively; $[$...$]_L$ denotes the stack of $L$ layers. 
 The output of the final layer $\mathbf{H}^L$ is returned as the representation of the input sentence.

\section{Approach}
The vanilla Transformer processes disperse words in a flat and uniform way, which makes it difficult to represent words in terms of their syntactic guidance~\cite{li2020transformer}.
Prior work has shown that decomposing the levels of granularity (phrases or templates) has produced substantial gains in paraphrase generation~\cite{li2019dnpg}, suggesting the possibility of further improvement from finer-grained modeling of granularity~\cite{hao2019multi}.

Motivated by the benefit of explicitly denoting word granularity, we propose GA-attention, a new self-attention block which automatically decomposes fine-grained granularity and learns granularity-aware sentence representations. We integrate GA-attention into the vanilla Transformer to generate granularity-aware paraphrases.

\subsection{Granularity-Aware Self-Attention}
\label{sec:approch}

\begin{figure*}[tb]
			\centering
			\subfloat[Architecture of the GA-attention network. The red dotted area represents the extensions against the original attention network.]{
				\includegraphics[width=0.3\textwidth]{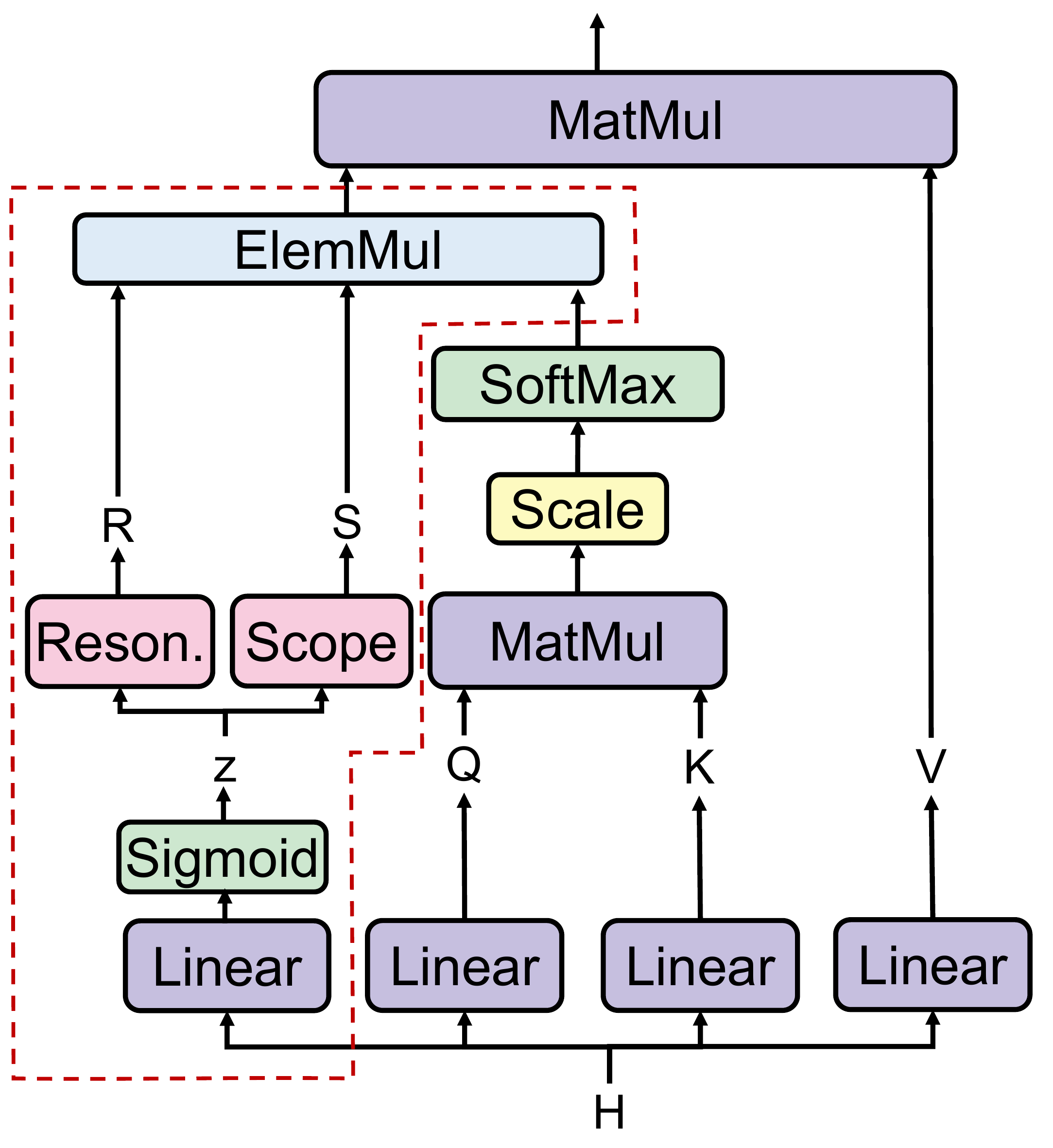}
				\label{fig:ga-attn}} 
			\space{      }\space{}\space{}\space{}
			\subfloat[Architecture of \approach which seamlessly replaces the multi-head attention with our multi-head GA-attention. Red arrows denote additional z flows in GA-attention.]{
				\includegraphics[width=0.48\textwidth]{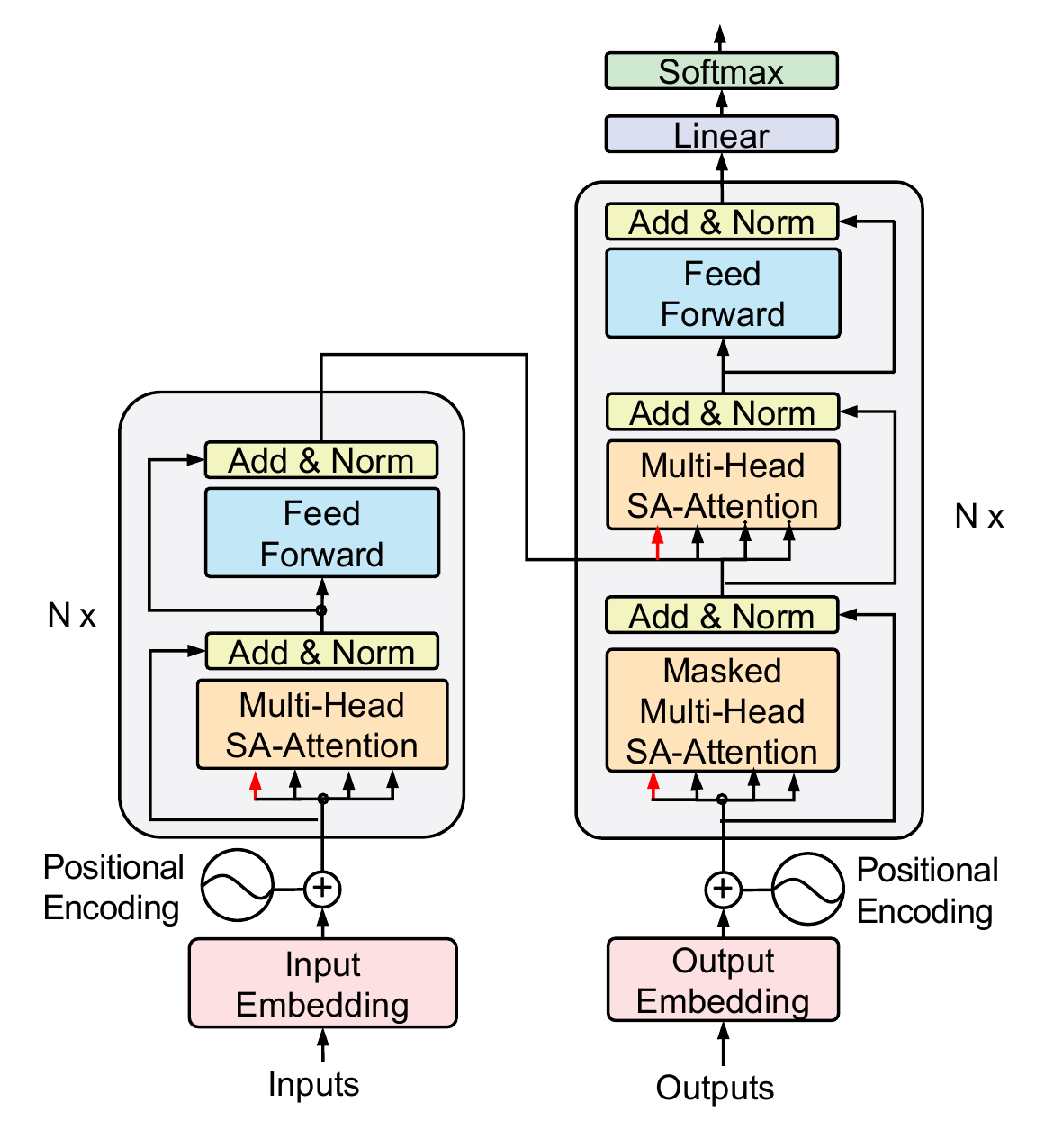}
				\label{fig:ga-transformer}}
			\caption{The architectures of GA-Attention and \approach.}
			\label{fig:arch}
		\end{figure*}

Unlike DNPG, which classifies each word into either phrase or sentence levels~\cite{li2019dnpg}, \approach aims to assign a soft classification of granularity for each word, yielding finer-grained decomposition of granularity for a sentence. 
For this purpose, we extend the vanilla self-attention with 1) a granularity head which estimates a continuous granularity level for each token, and 2) two new attention masks which bake the granularity into attentions for learning granularity-aware sentence representations. The overall architecture is illustrated in Figure~\ref{fig:ga-attn}.

\noindent\textbf{Granularity Head}
For a sequence of input tokens that is encoded as hidden states~$\mathbf{H}$ = [$\mathbf{h}_1,\ldots, \mathbf{h}_N$], the granularity head estimates a continuous granularity vector $\mathbf{z}=[z_1,\ldots,z_N]\in[0,1]^N$, where $z_i\in[0,1]$ measures the extent of token~$i$ belonging to details: a $z_i$ that is close to 1 indicates that the token at position~$i$ tends to be a detailed word (i.e., in the phrase level), while a $z_i$ approaching 0 indicates that token~$i$ tends to be a template word (i.e., in the sentence level). Specifically in the self-attention networks, the granularity for hidden states in layer~$l$ can be estimated as:
\begin{equation}
    \mathbf{z}^l=\mathrm{sigmoid}(\mathbf{W}^G\mathbf{H}^{l-1}), l=2,...,L
\end{equation}
where $\mathbf{W}^G$ represents the training parameters; $\mathbf{H}^{l-1}$ denotes the hidden states of layer $l$-$1$.

Having estimated the granularity for input tokens, we want to effectively incorporate the granularity into attentions to control the learning of sentence representations. To this end, we propose two new attention masks, namely, the granularity resonance mask and the granularity scope mask, 
Our idea is to adjust the original attention weights using the two proposed masks. 

\noindent\textbf{Granularity Resonance Mask }
We first introduce the \emph{granularity resonance mask} where ``resonance'' is analogy to an assumption of token correlations: sentence-level tokens attend more to sentence-level tokens, whereas phrase-level tokens attend more to phrase-level tokens~\cite{li2019dnpg}.
In this sense, the term \emph{granularity resonance} refers to the correlation between two tokens in terms of their levels of granularity. 

Let $z_i$ and $z_j$ denote the granularity of tokens $i$ and $j$ respectively. In the binary case where $z_i\in\{0,1\}$, their correlation in terms of granularity can be formulated as:
\begin{equation}
  \mathbf{C}_{ij} = 
  \begin{cases}
     1, & \mbox{if } z_i = z_j \\
     0, & \text{otherwise}
  \end{cases}
  \label{eq:reson:discrete}
\end{equation}
where $\mathbf{C}_{ij}$ represents the regularization coefficient to the original attention weight~$\mathbf{A}_{ij}$ in terms of granularity correlation. 
Such a discrete measure of resonance is limited in modeling token correlation as both $z_i$ and $\mathbf{C}_{ij}$ are binary variables. 
To improve the capacity of the granularity-aware attention, we generalize the computation of $\mathbf{C}_{ij}$ (Equation~\ref{eq:reson:discrete}) to a continuous function, that is,
\begin{equation}\label{eq:reson:continu}
    \begin{split}
       \mathbf{C}_{ij} = & (1 - z_i) \times \text{max}(0, 1-(z_i+z_j)) \\
                        & + z_i \times \text{min}(1, 1-z_i+z_j)
    \end{split}
\end{equation}
where $z_i$ $\in$ $[0, 1]$ is a continuous value; $z_i$ and 1-$z_i$ control the extent of token $i$ being in the word level or the sentence level, respectively. 
Equation~\ref{eq:reson:continu} provides a smooth measure of the correlation between token $i$ and $j$. 

\noindent\textbf{Granularity Scope Mask }
We further define the \textit{granularity scope mask} where \emph{granularity scope} measures the scope of attention according to the granularity level. This is based on the local attention assumption of phrases~\cite{li2019dnpg}:
a phrase-level token tends to attend to surrounding tokens while a sentence-level token can attend to other tokens evenly with any distance. 
In that sense, phrasal tokens (with a large $z_i$) have a relatively smaller attention scope compared to sentence-level words (with a small $z_i$).
For a given sequence of hidden states~$\mathbf{h}_1,\ldots,\mathbf{h}_N$ with $N$ words, the granularity scope for position~$i$ attending to position $j$ can be defined as: 
\begin{equation}
     \mathbf{S}_{ij} = 
     \begin{cases}
        1 & \text{if } |i-j|<(N-\epsilon)^{(1-z_i)}+\epsilon \\
        0 & \text{otherwise}
     \end{cases}
     \label{eq:scope:discrete}
\end{equation}
where $\mathbf{S}_{ij}\in\{0,1\}$ denotes the penalty for the original attention weight~$\mathbf{A}_{ij}$ in terms of granularity scope. $\epsilon$ denotes the maximum distance that a phrasal word attends to. We set $\epsilon$ to 2 according to a similar configuration in \cite{li2019dnpg}.
This equation can be intuitively interpreted as the receptive fields for different levels of granularity~\cite{li2019dnpg}: a phrase-level token~$i$~($z_i$=1) attends only to the adjacent $n$ ($n$=3) words, whereas a sentence-level token~$i$~($z_i$=0) can attend to positions with any distances.

Similarly, we generalize Equation~\ref{eq:scope:discrete} to a continuous function:
\begin{equation}
     \mathbf{S}_{ij}=\text{max}(0, \text{min}(1, (N-\epsilon)^{(1-z_i)}+\epsilon-|i-j|))
\end{equation}

 Using the two granularity-based attention masks, we derive the granularity-aware self-attention as an adjustment of the original attention weights, namely,
 
  \begin{equation}
   \begin{split}
     &\mbox{GASelfAttn}(\mathbf{H})= \tilde{\bf A}^T\mathbf{V} \\
     &\tilde{\bf A} = \mathbf{A}\odot\mathbf{C}\odot\mathbf{S}
   \end{split}
  \end{equation}
  where $\mathbf{A}$ and $\tilde{\bf A}$ denote the original and adjusted attention weights respectively; $\odot$ stands for the element-wise multiplication; 
  GASelfAttn(.) represents the proposed granularity-aware self-attention function.


\subsection{C-DNPG: Transformer with Granularity Aware Attention}

Based on the proposed GA-Attention, we propose \approach, which integrates GA-attention into Transformer in order to better generate paraphrases at fine-grained levels of granularity. Figure~\ref{fig:ga-transformer} illustrates the overall architecture of our model. Compared to the vanilla Transformer, \approach simply replaces the self-attention layers in both the encoder and the decoder with the proposed GA-attention network. 
Similar to the vanilla Transformer, we perform the GA-attention function for multiple heads in parallel and concatenate the multi-head representations to yield the final representation. 
The procedure for the \approach (Transformer with GA-Attention) can be summarized as:
\begin{equation}
  \begin{bmatrix}
       \mathbf{\bar H}^l = \mathrm{LN}(\mathrm{GASelfAttn}(\mathbf{H}^{l-1}) + \mathbf{H}^{l-1}) \\
       \mathbf{H}^l = \mathrm{LN}(\mathrm{FFN}^l(\mathbf{\bar H}^l) + \mathbf{\bar H}^l)
  \end{bmatrix}_L
\end{equation}
\noindent where LN denotes layer normalization~\cite{ba2016layer}. 

\section{Experiments}

\begin{table*}
\centering
\small
\begin{tabular}{l@{}c@{}c@{}c@{}c@{}c@{}c@{}c@{}c@{}c@{}c@{}c@{}}
\toprule
      &   \multicolumn{5}{c}{Quora}      & \;\;\;  &  \multicolumn{5}{c}{Twitter URL} \\
\cline{2-6}\cline{8-12}
Model  & \;iBLEU\; & BLEU-2\;  & BLEU-4\; & ROUGE-L\; &  METEOR & & iBLEU\; & BLEU-2\;& BLEU-4\; & ROUGE-L\; & METEOR \\ 
\hline
ResidualLSTM
&  20.45  & 40.71  &  26.20 & 36.19 & 32.67 &  & 20.29 & 36.75 & 25.92 & 32.47 & 29.44 \\ 
 Pointer-generator
 &  22.65  & 43.82 & 28.80 & 42.36 & 40.87 & & 25.60 & 44.50 & 32.40 & 38.48 & 36.48\\ 
 Transformer
 &  21.14  & 37.97 & 26.88 & 40.14 & 38.21 &  & 24.44 & 44.45 & 31.12 & 31.97 & 32.49 \\
 Transformer+Copy\;  & 22.90 & 44.42 & 28.94  & 37.60 & 38.34 &  & 27.07 & 48.44 & 34.35 & 38.37 & 38.19 \\
 DNPG
 & 24.55 & 47.72 &  31.01 &  42.37 &  42.12&  & 25.92 & 46.36 & 32.91 & 36.77 & 36.28 \\
 FSET
 & - & \bf 51.03 & 33.46  & - & 38.57  &  & -  & 46.35  & 34.62 & - & 31.67  \\
\hline
 C-DNPG (R)    & \bf 26.94& 47.58 &\bf 34.05 & 46.17 & 44.75 & & 27.96 & 49.98 & 35.80 & 38.67& 39.39\\
 C-DNPG (S)    & 26.68 & 47.48 & 33.93 & \bf 46.22 &\bf 46.66 & & 28.19 & 49.10 & 35.95 & 38.89 & 39.06 \\
 C-DNPG (R$\odot$S)  & 25.96 & 46.25 & 33.02 & 44.64 & 44.25 & &\bf 30.25 & 49.00 &\bf 38.58 &\bf 41.60 &\bf 41.71\\
 C-DNPG (R+S)  & 26.66 & 50.96 & 33.69  & 44.45 & 43.33 &    &  28.73 &\bf 50.49 & 36.61 & 39.80 & 40.42\\ 
\bottomrule
\end{tabular}
\caption{ Results of paraphrase generation on two benchmarks. R stands for the resonance mask while S stands for the scope mask; R$\times$S stands for the combination of two masks through element-wise multiplication; R+S means we average the two masks for adjusting the original attention weights. We note that the results of baseline models are stronger than those reported in the DNPG paper, probably due to the BERT tokenizer we have utilized in our experiments. The pointer-generator outperforms the vanilla Transformer, as is consistent to the DNPG paper.}
\label{table:result:quora}
\end{table*}

We evaluate our approach by experimenting on two widely used datasets, including the Quora question pairs and the Twitter URLs. We will introduce the common experimental setup and the empirical results.

\noindent \textbf{Implementation Details}
We implemented our approach on top of the Huggingface PyTorch Transformer~\cite{Wolf2019HuggingFacesTS}. 
For a fair comparison, we followed the hyperparameter settings in related works~\cite{li2019dnpg} for the Transformer. 
Both encoder and decoder consist of 3 transformer layers, have a hidden size of 450, and contain 9 attention heads (L=3, H=450, A=9). Following previous work~\cite{li2019dnpg,kazemnejad2020paraphrase}, we truncate sentences to 20 tokens. We utilize the pre-trained tokenizer by huggingface\footnote{https://github.com/huggingface/tokenizers} (i.e., bert-base-uncased) for tokenization which has been common in NLP. During decoding, we employ beam search with a beam size of 8. 
All models were optimized with AdamW~\cite{loshchilov2018adamw}. The learning rate was varied under a linear schedule with warmup steps of 5,000 and the maximum learning rate of 5$e^{-5}$. The model was training for 100,000 batches until achieving the best validation loss. The experiments were repeated for 5 times and were reported with their average results.
All models were trained on a machine with NVIDIA Tesla V100 GPU allocated with a batch size of 32 samples. 

\noindent \textbf{Baseline Models}
We compare our approach with popular paraphrase generation methods including: 
(i) RedidualLSTM~\cite{prakash2016reslstm}: an LSTM sequence-to-sequence model using residuals between RNN layers;
(ii) PointerGenerator~\cite{see2017pointer}: an RNN sequence-to-sequence model using copy mechanism;
(iii) Transformer~\cite{vaswani2017attention}: the vanilla Transformer model;
(iv) Transformer+Copy: an enhanced Transformer with copy mechanism~\cite{copy};
and 
(v) DNPG~\cite{li2019dnpg}: a popular paraphrase generation model based on Transformer. DNPG extends Transformer by generating paraphrases at multiple levels of granularity such as the phrase level and the sentence level. The model is composed of two encoder-decoder pairs, which correspond to phrase-level and sentence-level paraphrasing, respectively. 
We use the default settings of the baseline models as reported in their papers.
(vi) FSET~\cite{kazemnejad2020paraphrase}: the state-of-the-art paraphrase generation model that retrieves a paraphrase pair similar to the input sentence from a pre-defined index, then editing it using the extracted relations between the retrieved pair of sentences. We directly report the performance from their original paper.

\noindent \textbf{Evaluation Metrics}
We perform automatic evaluation using five widely used metrics for text generation tasks, namely, BLEU~\cite{papineni2002bleu}, iBLEU~\cite{sun2012ibleu}, ROUGE-L~\cite{lin2004rouge} and METEOR~\cite{lavie2007meteor}. 
We compute both BLEU-2 and BLEU-4 scores in our experiments using the NLTK package\footnote{https://www.nltk.org/\_modules/nltk/translate/bleu\_score.html}. 
iBLEU~\citep{sun2012ibleu} penalizes BLEU by n-gram similarity between output and input. Hence, it is taken as the main metric for paraphrasing. 

\noindent \textbf{Datasets}
We conducted the experiments on two widely used benchmarks:
1) the Quora question pairs benchmark\footnote{https://www.kaggle.com/c/quora-question-pairs (NC)}, which contains 124K duplicate question pairs. The dataset was labeled by human annotators and has been widely used for paraphrase research~\citep{li2019dnpg,devlin2018bert}. 
We split the original data into train, validation, and test sets with proportions of 100K, 4K, and 20K, respectively. \\
2) the Twitter URL paraphrasing dataset\footnote{https://github.com/lanwuwei/Twitter-URL-Corpus} is also a widely used benchmark for evaluating paraphrase generation~\cite{li2018paraphrase,kazemnejad2020paraphrase}. The dataset contains two subsets which are manually and automatically labeled, respectively. Following \cite{li2018paraphrase}, we sample 110k instances from the automatically labeled subset as our training set and sample 5k and 1k instances from the manually annotated subset for the test and validation sets, respectively.

\section{Results and Analysis}

\subsection{Automatic Evaluation}
Table~\ref{table:result:quora} shows the results of various approaches on the two benchmarks. As the results indicate, \approach (with variants) achieves the best performance in terms of most automatic metrics, which suggests that our \approach is effective in performing multi-granularity paraphrasing. 

In particular, our approach outperforms DNPG, a multi-granularity Transformer based model with a significant margin. That means that by modeling more fine-grained granularity levels, \approach can control the generating process more precisely. Thanks to the granularity attention mechanisms, it is more flexible for the model to leverage syntactic guidance (e.g., recognizing templates and details) for paraphrase generation. 

It is interesting to note that either the resonance mask or the scope mask that we propose can achieve the best performance under specific settings. We hypothesize that there could be overlap between the two proposed attention masks in some cases. Therefore, combining them may amplify the extent of masking and hinder the ultimate performance. We also find that Transformer with a copy mechanism can outperform DNPG on the Twitter dataset. This might be due to the more noise in this dataset, which leads copy based models to be more effective.  

	\begin{figure} [h]
		\centering 
		\includegraphics[width=0.5\textwidth]{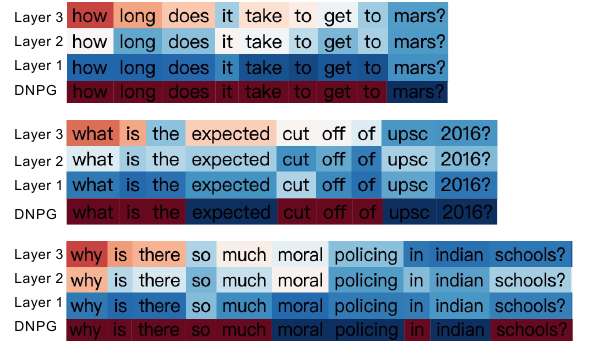} 
		\caption{Examples of multi-granularity extracted by \approach (Layer1-3) and DNPG (bottom) on the Quora dataset. Warmer colors represent higher levels of granularity (templates) while colder colors represent lower levels of granularity (details). We present granularity of all Transformer layers and compare the results with those of DNPG.}
		\label{fig:case}
	\end{figure}

\label{table:ablation:comp}

\subsection{Qualitative Analysis}

\begin{table*} [!htb]
\centering
\begin{tabular}{rl}
\toprule
\bf Sentence: & What is a good first programming language? \\
                             \hline
  \bf Transformer: & What is good? \\ 
  \bf DNPG: &  What is good for coding? \\ 
  \bf C-DNPG: & What are the best programming languages for beginners?\\
  \bf Human: & Whats a good and easy programming language to learn?  \\
\bottomrule
\end{tabular}
\quad\\
\quad\\
\begin{tabular}{rl}
\toprule
\bf Sentence: & What will the year 2100 be like? \\
                             \hline
  \bf Transformer: & What is likely to happen in the world? \\ 
  \bf DNPG: &  What are did today. year - year of unique year of country?\\ 
  \bf C-DNPG: & What will the world look like in 2100?  \\
  \bf Human: & What will the year 2099 be like? \\
\bottomrule
\end{tabular}
\caption{Sample paraphrases from multiple models with human reference.}
\label{table:case}
\end{table*}

To gain a more in-depth insight into the performance, we qualitatively analyze the interpretability of GA-attention. 
We visualize the output of the granularity head in each attention layer to verify how effectively GA-attention captures fine-grained granularity.
As shown in Figure~\ref{fig:case}, GA-attention can successfully capture continuous linguistic structures reflected as multiple levels of granularity (in the last layer). For example, it successfully yields four levels of templates: 1) \emph{what} \underline{\space\space\space\space}, 2) \emph{what is} \underline{\space\space\space\space}, and 3) \emph{what is the expected} \underline{\space\space\space\space} 4) \emph{what is the expected cut off of} \underline{\space\space\space\space} according to Example 1 in Figure~\ref{fig:case}. In contrast, DNPG can decompose only two levels of granularity for each sentence. 
This means that \approach can successfully distinguish templates and detailed words for each sentence, thus generating more fine-grained paraphrases.

Another interesting observation is that the continuous granularity is not extracted at once, instead, it is gradually summarized through transformer layers. This indicates that the proposed granularity-aware extensions blend naturally with attention networks in learning \emph{coarse-to-fine} representations~\cite{jawahar2019does}.

Overall, the results suggest that the proposed GA-attention naturally extends vanilla attention networks and enhances text representations with fine-grained granularity modeling.

\subsection{Case Study}

Table~\ref{table:case} presents two sample paraphrases generated by different models in the Quora test set.
As the samples indicate, \approach generates more coherent and fluent paraphrases than other models, which is consistent with the results of the automatic and human evaluation.
According to the first sample, \approach produces a more relevant and human-like paraphrase. For example, \approach successfully paraphrases the word ``first'' as ``\emph{beginners}''. 
The second sample shows more clear strength of \approach which generates a paraphrase that is even better than the ground-truth question asked by human (e.g., the year ``2100'' is mistakenly paraphrased as ``2099'').  

\begin{table*} [htb]
\setlength{\belowcaptionskip}{0pt}
\centering
\begin{tabular}{lcccccccc}
\toprule 
{\textbf{Comparison}} &
    \multicolumn{4}{c}{{Relevance}}  & \multicolumn{4}{c}{{Fluency}} \\
\cline{2-9}
 & \textbf{Win} & \textbf{Tie} & \textbf{Loss} & Kappa &
    \textbf{Win} & \textbf{Tie} & \textbf{Loss} & Kappa\\
\midrule
Ours \textit{vs.} Transformer &  67.8\% & 14.3\% & 17.8\% & 0.156 & 67.5\% & 15.8\% & 16.7\% & 0.166 \\
Ours \textit{vs.} DNPG & 72.3\% & 12.8\% & 14.8\% & 0.171  & 68.1\% & 17.5\% & 14.3\% & 0.194\\
Ours \textit{vs.} Human &  43.7\%  & 10.5\%  & 45.8\% & 0.079 & 43.2\%  & 11.0\% & 45.8\% & 0.095\\
\bottomrule
\end{tabular}
\caption{Human evaluation on the test set of Quora.}
\label{table:results:human}
\end{table*}

\subsection{Human Evaluation}

Besides the automatic evaluation, we also perform a human study to assess the performance of our approach qualitatively. We compare our approach with two typical methods, namely DNPG~\cite{li2019dnpg} and the vanilla Transformer~\cite{vaswani2017attention}. They represent the state-of-the-art decomposition based method and the backbone model that our model is built upon, respectively. We randomly selected 200 Quora questions from the test set. For each one of the questions, one paraphrase was generated for each model. Then, three annotators from the Amazon Mechanical Turk were asked to compare the generated paraphrases by two models (ours vs. a baseline model) blindly based on two criteria, relevance and fluency. Relevance means that the generated paraphrases are semantically equivalent to the original question. Fluency means the generated paraphrases are natural and fluent sentences. Table~\ref{table:results:human} presents the comparison results. As can be seen, our model significantly outperforms the other two methods in terms of the two criteria. Moreover, the Fleiss' kappa $\kappa$ shows fair agreement between annotators.

\begin{table}[h]
\centering
\begin{tabular}{lc}
\toprule
  Model & Time (hours) \\
 \toprule
  Transformer\; & 1.2 \\ 
  TransformerCopy & 1.2 \\ 
  DNPG &  3.6 \\ 
 \hline
  C-DNPG (ours) & 1.5  \\ 
\bottomrule
\end{tabular}
\caption{Training time (until model convergence) of various approaches on the Quora dataset.}
\label{table:time}
\end{table}

\subsection{Computational Efficiency}
As one of the key advantages of \approach, we finally evaluate the time efficiency of our approach. We used the same setup as described in the Setup section. As Table~\ref{table:time} shows, the granularity aware attention mechanism in \approach does not bring much additional computational cost to Transformers as opposed to the DNPG baseline approach, which indicates that GA-attention is lightweight to be integrated into Transformers.

\section{Discussion}
\subsection{Why baking the two masks can enhance the performance?}

Our idea is a generalization of the previous work DNPG~\cite{li2019dnpg}. In that paper, decomposing sentences into templates and details can improve the performance of paraphrasing because paraphrasing is usually generated by rewriting the sentence template while directly copying the detailed words. In that sense, template words pay more attention to template words while detailed words tend to pay attention to detailed words. Our paper generalizes this idea by extending the binary level (either 0 or 1) of granularity to a continuous range (between 0 and 1) of levels. This was implemented by the two proposed masks.

\section{Related Work}
This work is closely related to (1) multi-granularity paraphrase generation and (2) multi-granularity attention.

\noindent\textbf{Multi-Granularity Paraphrase Generation}. 

There has been an increasing interest in decomposing paraphrase generation into multiple levels of granularity such as word, phrase and sentence~\cite{li2019dnpg,wiseman2018learning,hao2019multi}. 
For example, \cite{li2019dnpg} present the decomposable neural paraphrase generator (DNPG). DNPG is a Transformer-based model that generates paraphrases at two levels of granularity in a disentangled way. The model is composed of two encoder-decoder pairs, corresponding to phrase-level and sentence-level, respectively. The difference between our \approach and DNPG is of three-fold: 1) \approach estimates a continuous granularity level of each token and hence supports a continuous modeling of hierarchical structures; 2) Compared to DNPG, the GA-attention in \approach can be naturally and seamlessly integrated into Transformer and is light-weighted in computation; and 3) While DNPG predicts granularity using a single fully connected layer, \approach gradually summarizes granularity through a stack of Transformer layers.

\noindent\textbf{Multi-Granularity Attention}.

Another important line of work relates to multi-granularity self-attention. 
\cite{hao2019multi} proposed multi-granularity self-attention (MG-SA): MG-SA combines multi-head self-attention and phrase modeling by trains attention heads to attend to phrases in either $n$-gram or syntactic formalism. 
\cite{nguyen2020tree} proposed a tree-structured attention network which encodes parse tree structures into self-attention at constant time complexity.
Despite similar names, our method differs from theirs greatly in principle and architecture. These two works rely on the existence of parsing trees, as opposed to GA-attention which infers latent structures from plain text. Furthermore, MG-SA only considers two-levels of granularity while GA-attention aims at continuous modeling of multiple granularity. 
\cite{liu2020text} proposed a hybrid neural architecture named MahNN which integrates RNN and ConvNet, each learning a different aspect of semantic from the linguistic structures. 
Like other related works, MahNN is based on a coarse-grained attention mechanism. The two levels of granularity it processes are represented by RNN and ConvNet, respectively. By contrast, GA-attention provides a fine-grained attention function extended from the vanilla attention mechanism. Therefore, GA-attention is a pure attention-based approach and can be naturally and seamlessly integrated into Transformers.




\section{Conclusion}
In this paper, we have proposed a novel paraphrase generation model named \approach for continuously decomposing sentences at different levels of granularity.
\approach extends the multi-head attention with a granularity head which neurally estimates continuous granularity level of each input token.
To efficiently incorporate granularity into attentions, we propose two novel attention masks, namely, granularity-resonance mask and granularity-scope mask, to adjust the original attention weights.
Results on two paraphrase generation benchmarks show that \approach remarkably outperforms baseline models in both quantitative and qualitative studies.
In future work, we will investigate the effect of \approach in pre-trained models and other NLP tasks.

\section*{Acknowledgments}
We would like to thank the anonymous reviewers for their constructive feedback. Xiaodong Gu was sponsored by NSFC No. 62102244, CCF-Tencent Open Research Fund (RAGR20220129), and CCF-Baidu Open Fund (NO.2021PP15002000).


\bibliography{references}
\bibliographystyle{acl_natbib}

\end{document}